\documentclass[10pt,twocolumn,letterpaper]{article}

\usepackage[pagenumbers]{cvpr} 

\usepackage{multirow}
\usepackage{caption}
\usepackage{times}
\usepackage{epsfig}
\usepackage{makecell}
\usepackage{graphicx}
\usepackage{amsmath}
\usepackage{amssymb}
\usepackage{booktabs}
\usepackage{xcolor}

\newcommand\hr[1]{\textcolor{red}{\textbf{#1}}}
\newcommand\hb[1]{\textcolor{blue}{\textbf{#1}}}
\usepackage[pagebackref,breaklinks,colorlinks]{hyperref}

\usepackage[capitalize]{cleveref}
\crefname{section}{Sec.}{Secs.}
\Crefname{section}{Section}{Sections}
\Crefname{table}{Table}{Tables}
\crefname{table}{Tab.}{Tabs.}

\def\cvprPaperID{10266} 
\def\confName{CVPR}
\def\confYear{2025}

\begin{document}

\title{Time Step Generating: A Universal Synthesized Deepfake Image Detector}

\author{Ziyue Zeng$^*$, Haoyuan Liu, Dingjie Peng, Luoxu Jin, Hiroshi Watanabe\\
Waseda University\\
Nishiwaseda, Shinjuku City, Tokyo 169-0051\\
{\tt\small zengziyue@fuji.waseda.jp}
}

\maketitle

\begin{abstract}
   Currently, high-fidelity text-to-image models are developed in an accelerating pace. Among them, Diffusion Models have led to a remarkable improvement in the quality of image generation, making it vary challenging to distinguish between real and synthesized images. It simultaneously raises serious concerns regarding privacy and security. Some methods are proposed to distinguish the  diffusion model generated images through reconstructing. However, the inversion and denoising processes are time-consuming and heavily reliant on the pre-trained generative model. Consequently, if the pre-trained generative model meet the problem of out-of-domain, the detection performance declines. To address this issue, we propose a universal synthetic image detector  \textbf{T}ime \textbf{S}tep \textbf{G}enerating (\textbf{TSG}), which does not rely on pre-trained models' reconstructing ability, specific datasets, or sampling algorithms. Our method utilizes a pre-trained diffusion model's network as a feature extractor to capture fine-grained details, focusing on the subtle differences between real and synthetic images. By controlling the time step t of the network input, we can effectively extract these distinguishing detail features. Then, those features can be passed through a classifier (\textit{i.e.} Resnet), which efficiently detects whether an image is synthetic or real. We test the proposed \textbf{TSG} on the large-scale GenImage benchmark and it achieves significant improvements in both accuracy and generalizability. The code and dataset are available at: https://github.com/NuayHL/TimeStepGenerating
\end{abstract}
\section{Introduction}

\footnotetext[1]{$^*$Ziyue Zeng is the first and corresponding author.}

Recently, diffusion models have achieved state-of-the-art performance in the field of image generation. The Denoising Diffusion Probabilistic Models (DDPMs)\cite{ddpm} have introduced a new method for high-quality image generation and have been widely researched. Improvements to diffusion models have focused on multiple aspects, such as accelerating sampling \cite{ddim,timesteptuner,attentionEfficiency}, innovate the backbone network\cite{diffussm,zhang2024improving}, improved model framework\cite{ADM,sohl2015deep,sd} and optimizing training strategies\cite{nichol2021improved,ho2022classifier}. Diffusion models have also been investigated for various downstream tasks, including video generation \cite{videodiff}, controllable image synthesis \cite{controlnetplus,peng2024controlnext}, and image editing \cite{imagic,brooks2023instructpix2pix}. The proliferation of diffusion model-based technologies in everyday life has raised significant concerns\cite{juefei2022countering} regarding privacy, the dissemination of misleading information, and copyright infringement. Therefore, it is imperative to develop a method to detecting generated images to ensure the integrity of a trustworthy social environment.

\begin{figure}[t]
\begin{center}
   \includegraphics[width=0.88\linewidth, height=7cm]{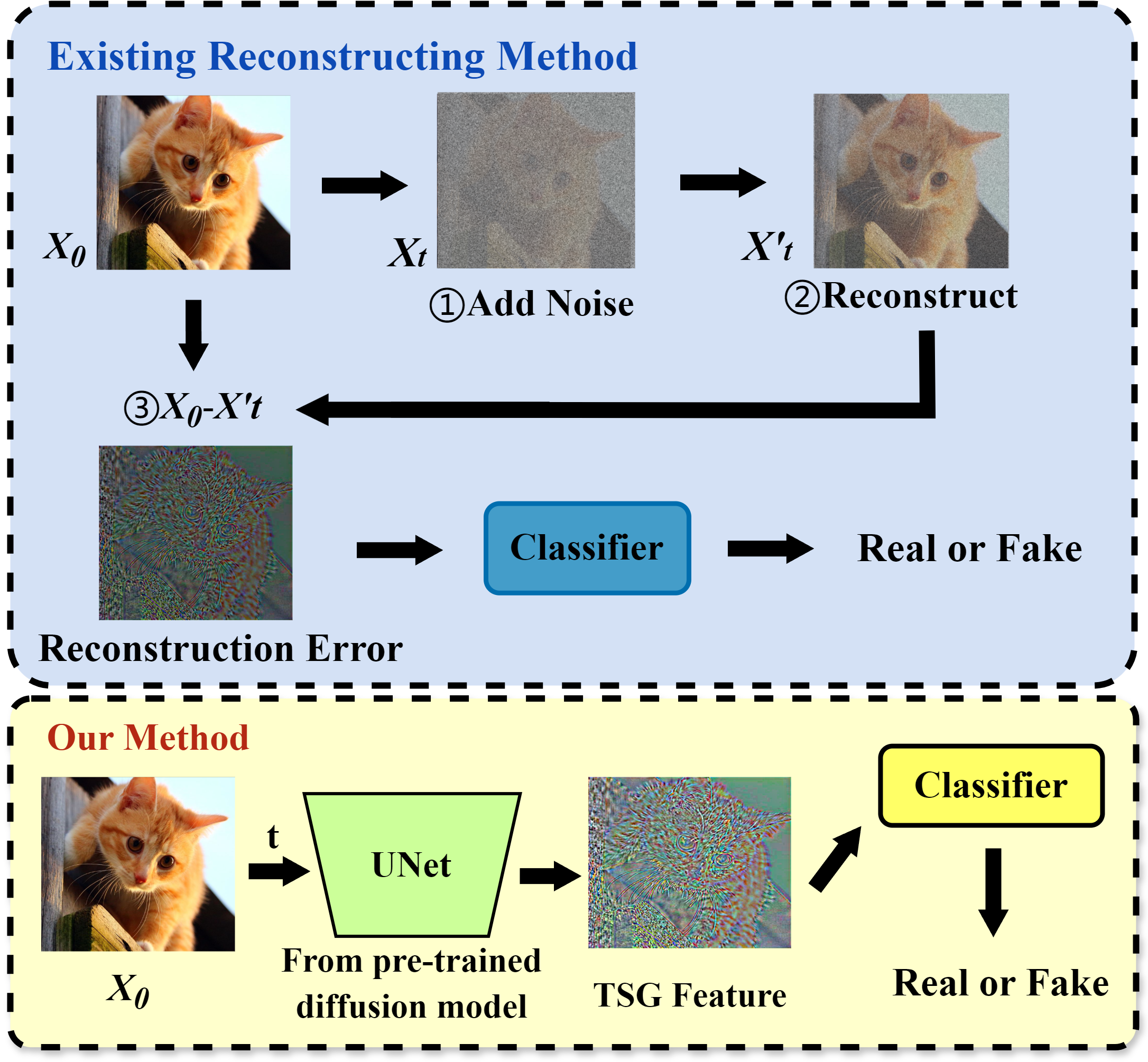}
\end{center}
   \caption{Overview of the reconstructing based method and our method.
   In the reconstructing method, $X_0$ is the original picture and we add noise to the original image through one or several inverse processes to get $X_t$.  $X'_t$ is the image obtained after denoising $X_t$. In the proposed method TSG, we first fix the timestep \(t\) and extract features using a pretrained U-Net neural network from a diffusion model. Then, these features are fed into a classification network for prediction.}
\label{map}
\end{figure}

The diffusion model differs substantially from previous methods in image generation, and existing research techniques struggle to accurately identify images produced by diffusion models. To develop a universal detector for images generated by diffusion models, experts have achieved promising results by investigating the principles underlying diffusion models. To address these issues, Diffusion Reconstruction Error (DIRE)\cite{dire} was proposed, based on the hypothesis that authentic images present greater challenges for diffusion models to reconstruct accurately. Therefore, the reconstruction error between the generated image and the original image can serve as an effective classification feature. To enhance the generation speed of DIRE, LaRE$^2$\cite{lare} was introduced, leveraging one-step reconstruction error within the latent space as a central approach. However, DIRE has two main shortcomings: its time-consuming reconstruction process and its reliance on the performance of the reconstruction model, both of which have yet to be fully addressed.

In response to these concerns, we propose Time Step Generating (TSG). Compared to the reconstruction processes of single-step inversion and denoising, TGS simplifies the approach further. In the process of TSG, we consider the following points: Firstly, based on the premise that real images are more challenging to reconstruct than generated images, we focus on the most crucial step in the reconstruction process: the estimation of the noise \(\epsilon\). Taking DDPM as an example, considering the time step \(t\) and varying details contained in the input image, the U-Net estimates the noise presented in the image at that specific moment. Likewise, the pre-trained U-Net's ability to estimate noise also differs between real images and generated images. Secondly, from the perspective of Score-Based Diffusion model, the denoising process is regarded as moving toward directions of higher probability density in the sample space. Consequently, for real images, which are already situated at the highest probability density point, the gradient at this point differs from the estimated gradient of the generated images. Based on the above two inferences, a pre-trained U-Net is employed to directly extract features from the images to be classified. The schematic diagram of TSG is shown in Figure~\ref{map}.

Generally, in TSG, we first use a neural network to extract features from the images, and then feed these features into a classification network for identification. However, unlike previous work, the feature extraction network we use comes from a pre-trained unconditional diffusion model. Theoretically, any pre-trained diffusion model can serve as the feature extractor for TSG, as long as it has sufficient capability for generating detailed features. Thus, our method can handle datasets with arbitrary content and various generators, significantly enhancing versatility and model application speed.

To evaluate our method, we use the GenImage benchmark\cite{zhu2024genimage}. It consists of 8 subsets generated by different kinds of generators,  each containing about 300,000 image samples. Compared to DIRE, our method is 10 times faster and compared to LaRE$^2$ we achieves  substantial improvement in accuracy. 
The contributions of our work are as follow:
\begin{enumerate}
    \item \textbf{New feature extraction paradigm:} Unlike reconstruction based feature extractors, our method further decomposes the single-step reconstruction process based on the principles of diffusion models, focusing on the step that best distinguishes real images from generated ones: noise prediction.
    \item \textbf{Greater generalizability:} Our method requires no additional training beyond the classifier. The generalizability of the TSG is greatly enhanced compared to reconstruction models, as it does not require consideration of whether it can reconstruct for specific datasets or types of objects.
    \item \textbf{Better performance:} We conducted extensive experiments that demonstrate a significant improvement in the accuracy of our method compared to previous baseline.
\end{enumerate}

\section{Related Works}

\subsection{Image Generation}

Generative Adversarial Networks\cite{gan} (GANs) and Variational Autoencoders\cite{vae} (VAEs) have been pioneers in the field of image generation. However, they are limited by the quality of generated images and the stability of training. Diffusion models represent a newer approach to image generation, particularly notable for their stable training and quality in creating detailed and diverse images. These models work by defining a reverse process that progressively refines images from pure noise to detailed data, guided by probabilistic modeling. This approach captures complex data distributions and avoids some common challenges seen in GANs, like mode collapse.
Denoising Diffusion Probabilistic Models\cite{ddpm} (DDPM) are based on a Markov chain of steps that iteratively add noise to data and then learn to reverse this noise. The forward process involves gradually corrupting the data with Gaussian noise until it becomes unrecognizable. In contrast, the reverse process learns to reconstruct the image from noise, resulting in high-quality generation. This method is particularly effective for generating diverse and high-resolution images by modeling the gradual improvement of details over a series of steps. Moreover, recent work such as Stable Diffusion\cite{sd}, PixArt-$\alpha$\cite{pixel} have reached state-of-the-art status in text-to-image generation.

\subsection{Generated Image Detection}

In the past few years, research on detecting generated images has primarily centered on images produced by GAN-based generation methods\cite{wang2020cnn,chai2020makes}. Detection approaches for GAN-generated images have largely relied on feature detection, which attempt to identify subtle artifacts and inconsistencies specific to GANs. These methods leverage Convolutional Neural Networks (CNNs) to analyze and classify visual features, enabling the identification of synthetic images by learning unique patterns that distinguish GAN-generated content from real images. Some works focus on detecting fake faces \cite{face1,liang2024poisoned,zhang2022improving}, while others focus on general models\cite{wissmann2024whodunit,he2024rigid}. However, with the rise of newer models like diffusion-based image generation,these CNN-based feature detection methods have shown limitations. To address these challenges, recent research has explored innovative detection approaches tailored to the unique characteristics of diffusion models. Wu et al.\cite{wu2023} introduced a method that leverages a CLIP-based model for detection\cite{clip}, utilizing the feature representations of CLIP to better distinguish between real and diffusion-generated images. Wang et al.\cite{dire} proposed the Diffusion Reconstruction Error (DIRE) method which exploits the error in real image reconstruction for image detection. Cazenavette et al.\cite{Cazenavette_2024} further develop synthetic image detection by utilizing inversion feature maps for classification.
Luo et al.\cite{lare} perform an effective detection method compared to the DIRE method, while Tan et al.\cite{Tan_2023} propose a gradient-based detection approach. Inspired by this previous work, we propose the Time Step Generating (TSG), which improves both accuracy and speed.

\section{Methods}
We first provide background information on Score-Based Diffusion Model and DDPM in Section~\ref{sec:prelim}. Then, we introduce our proposed TSG feature extraction method  in Section~\ref{sec:tsg}.
\subsection{Preliminaries}
\label{sec:prelim}
\textbf{Score-Based Diffusion Model} constructs a diffusion process $\{x_t\}_{t=0}^T$ indexed by a continuous time horizon $[0, T]$, which gradually transform a data distribution $q_0(x_0)$ into a noise distribution $q_T(x_T)$ using a stochastic differential equation (SDE). The forward process can be formulated as: 
\begin{equation}
\label{eq:score-forward}
\mathrm{d}x=\boldsymbol{f}_t(x)\mathrm{d}t+g(t)\mathrm{d}\boldsymbol{w},
\end{equation}
where $\boldsymbol{w}$ is standard Wiener process. $\boldsymbol{f}_t(\cdot)$ and $g(t)$ denote the drift coefficient and diffusion coefficient, respectively. The reverse process relies exclusively on the time-dependent gradient field (or score) of the perturbed data distribution as:
\begin{equation}
\label{eq:score-solu}
    \mathrm{d}x = \left[\boldsymbol{f}_t(x)-g(t)^2\nabla_{x}\log q_t(x)\right]\mathrm{d}t + g(t)\boldsymbol{\bar{w}},
\end{equation}
where $\boldsymbol{\bar{w}}$ and $\mathrm{d}t$ denote the standard Wiener process in the reverse-time direction and an infinitesimal negative time step, respectively. Score-based models estimate this gradient field $\nabla_{x_t} \log q_t(x_t)$ by training a neural network $s_\theta(x_t, t)$, with score matching losses as the objective:
\begin{equation}
\label{eq:score-dsm}
\begin{aligned}
&\mathcal{J}_{\text{DSM}}(\theta) :=\\
 &\mathbf{E}_{q_0(x_0) q_{t}(x_t|x_0)}\left[{||\nabla_{x_t} \log q_{t}(x_t|x_0) - s_\theta(x_t, t)||}_2^2\right].  
\end{aligned}
\end{equation}
The score-based model $s_\theta(x_t, t)$ can be incorporated into (\ref{eq:score-solu}) once the it has finished training. Subsequently, samples are generated by numerically solving this reverse process SDE, which retraces the forward diffusion in (\ref{eq:score-forward}) in reverse time, ultimately producing an approximate data sample.

\textbf{Denoising Diffusion Probabilistic Model} is intuitively composed of a forward process and a reverse process. In the diffusion forward process, let $x_0$ be the original image selected from the dataset, random Gaussian noise is sampled and gradually add to $x_0$, denoted as: 
\begin{equation}
    q(x_t|x_{t-1})=\mathcal{N}(x_t;\sqrt{\frac{\alpha_t}{\alpha_{t-1}}}x_{t-1},(1-\frac{\alpha_t}{\alpha_{t-1}}\textbf{I})),
\end{equation}
where $x_t$ denotes the images in the process of adding noise, $t$ and $\alpha_t$ are two pre-defined sequence of hyperparameters. An important corollary is that we can directly add noise in a single step to obtain a noisy image at any given time step:
\begin{equation}
    q(x-t|x_0)=\mathcal{N}(x_t;\sqrt{\alpha_t}x_0,(1-\alpha_t)\textbf{I}).
\end{equation}
The reverse process is also defined as a series of Markov processes following normal distribution:
\begin{equation}
    p_\theta(x_{t-1}|x_t)=\mathcal{N}(x_{t-1};\mu_\theta(x_t,t),\Sigma_\theta(x_t,t)),
\end{equation}
where $p_\theta$ can be indirectly predicted by a neural network $\epsilon_\theta$. In the training process, $\epsilon_\theta$ take $x_t$ and time step $t$ as input to predict the added noise $\epsilon$, which is defined as:
\begin{equation}
    \mathcal{L}_\theta(x_0,t)=||\epsilon-\epsilon_\theta(\sqrt{\bar{\alpha}_t}x_0+\sqrt{1-\bar{\alpha}}_t\epsilon,t)||_2^2.
\end{equation}

\subsection{Time Step Generating}
\label{sec:tsg}

\begin{figure}[t]
\begin{center}
   \includegraphics[width=0.75\linewidth]{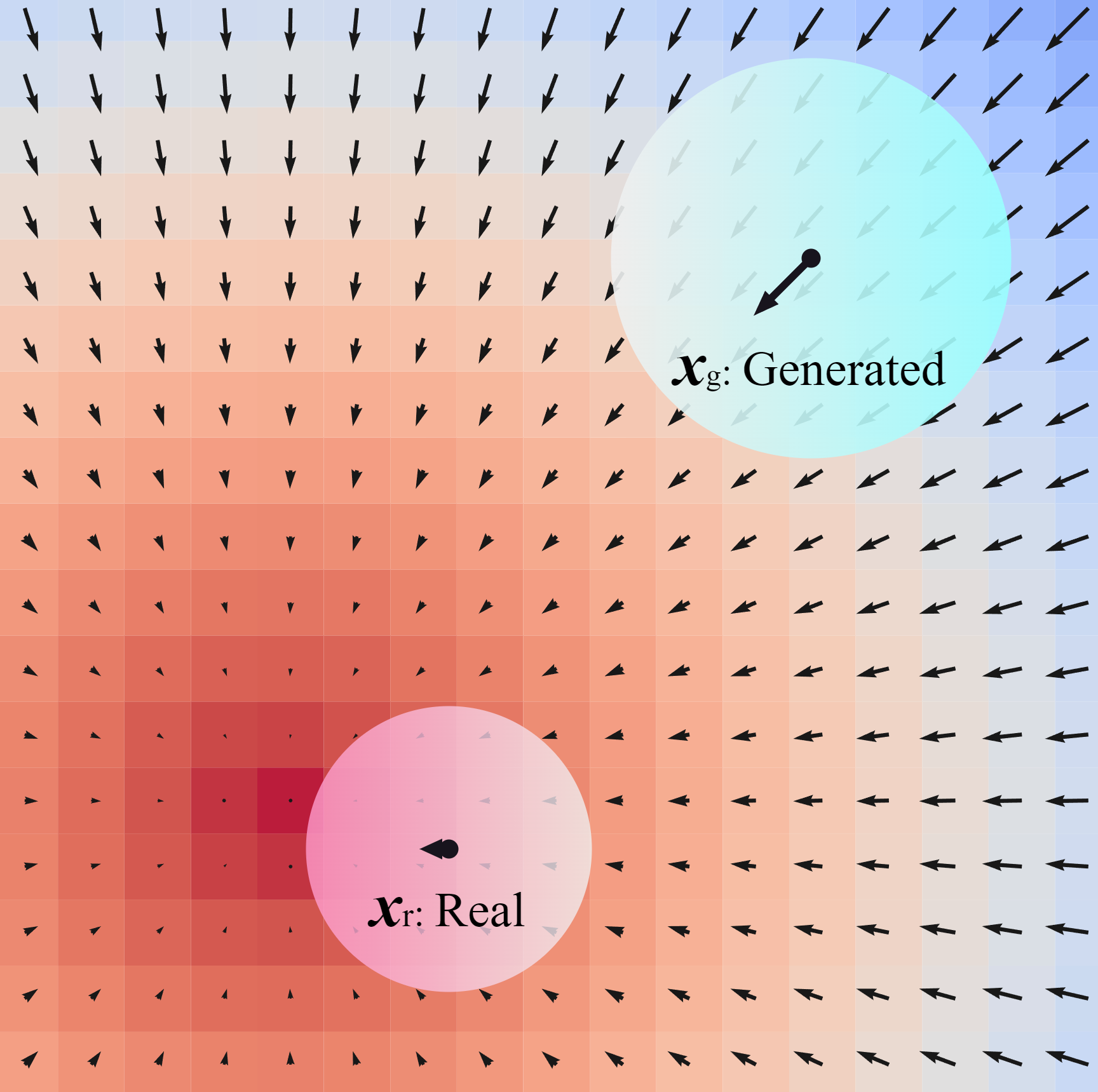}
\end{center}
   \caption{Explain the differences between real and generated samples from the perspective of scores. $x_r$ is the real image's distribution and $x_g$ represents the distribution of generated images. We take the center point of the distribution as an example, the arrow at the center of the distribution represents the estimated score at this point.}
\label{fig2}
\end{figure}

We first assume a reverse denoising process of a DDPM, the \(x_t\) near the completion of the reverse process already contains numerous details and is close to samples of images generated by the generative model. TSG feeds the original images, along with a timestamp close to the end of the generation process, into the U-Net within a pre-trained diffusion model. From the DDPM perspective, setting \(t\) close to 0 in the reverse process refines details for the given image. Therefore, the noise predicted by the neural network will naturally contain the detailed information of the input image. On the other hand, Figure~\ref{fig2} illustrates our TSG from the perspective of Score-Based Diffusion Model, where the two circles represent the distributions of real and generated images, respectively. Real images' distribution $x_r$ is originally located in areas of high probability density, so the predicted score at this point leads to significant differences from that of the generated image, providing distinct detailed features that can be detected by ResNet-50\cite{resnet}. If noise is added to the real images in the same way as in previous work. the distribution of real images will be significantly affected, and the original real-detail features are also disrupted. This may help explain why, in later experiments, reconstruction methods perform worse than TSG's one-step feature extraction.

The definition of TSG is as follow:
\begin{equation}
    F=\epsilon_\theta(I,t),
\end{equation}
where $I=x_r$ or $x_g$, and $F$ is the feature extracted by the U-Net. Then, we feed $F$ into the ResNet-50 for binary classification.

\begin{figure}[t]
\begin{center}
   \includegraphics[width=0.97\linewidth]{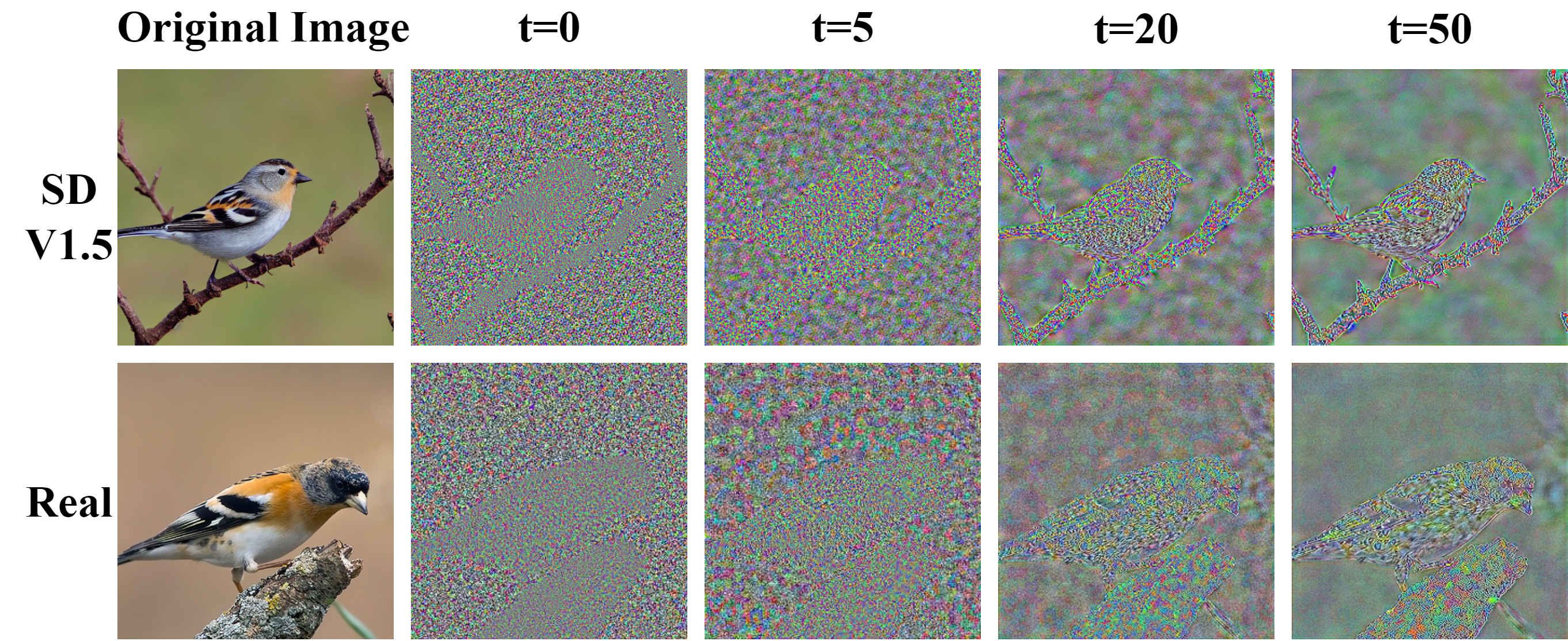} 
\end{center}
\caption{The feature images output by TSG under different conditions of \( t \).}
\label{fig3}
\end{figure}

Figure~\ref{fig3} shows some examples of the TSG images. We can see that as $t$ increases, the main object's outline in the feature map generated by TSG becomes progressively clearer, highlighting high-frequency information. Comparing the feature maps at \( t = 0 \) and \( t = 50 \), we can observe that at \( t = 0 \), the main part of the image shows more detailed noise. As \( t \) increases, the noise in both the main part and the background gradually decreases. Next, we will conduct experiments to perform a more detailed analysis of TSG.

\section{Experiments}
\subsection{Datasets and Pretrained Models}
We use the GenImage\cite{zhu2024genimage} dataset to test the detection accuracy of our proposed method and analyze changes brought by different time step $t$. In the GenImage dataset, we select the images from  5 different generative models: \textit{BigGAN}\cite{biggan}, \textit{VQDM}\cite{gu2022vector}, \textit{SD V1.5}\cite{sd}, \textit{ADM}\cite{ADM} and \textit{Wukong}\cite{gu2022wukong}, where the last four datasets are generated by diffusion models. Each subset of generated models contains approximately 330k images in total, and divided into training and validation sets. To facilitate comparison with previous work, we follow the dataset's original settings to divide the training and validation sets. 

In all experiments, we adopt the class-unconditional ImageNet diffusion model at resolution $256\times256$ which released with the paper \cite{ADM}.

\subsection{Implementation Details}
Our code is basically modified based on DIRE. During the feature extraction process, images are resized to $256\times256$ and input to the U-Net. Then, during the classifier training phase, we crop the images with a size of $224\times224$ and use ResNet-50\cite{resnet} as the base network. To study the generalization of our method, we selected images generated by five different generation methods as subsets and trained a classifier on each subset for subsequent classification experiments. The results and other experiments are presented below.

\subsection{Generalization on Generative Methods}

\begin{figure*}[!ht]
\begin{center}
   \includegraphics[width=0.9\linewidth]{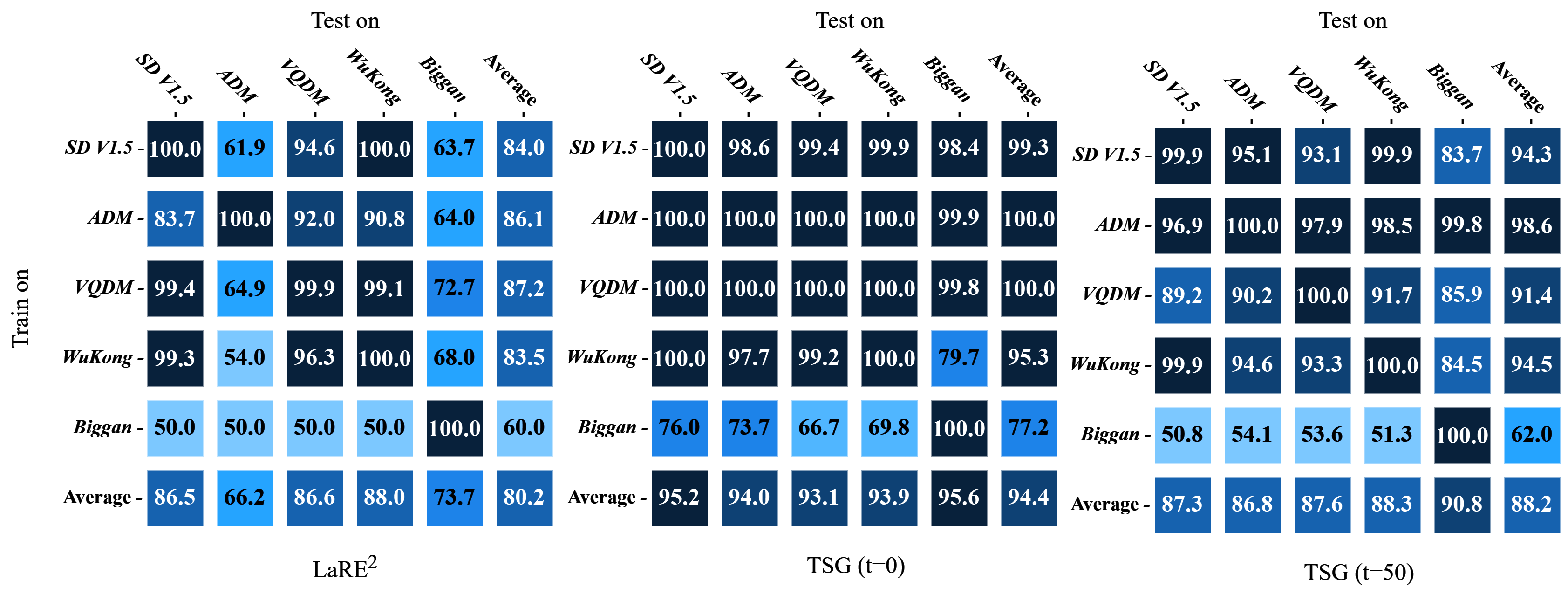}
\end{center}
   \caption{Cross validation results on various training and testing subsets of Genimage. For each generator, a model is trained and tested across all 5 generators. The matrix plot presents the accuracy of LaRE$^2$ and TSG, with TSG evaluated under two parameter settings: $t=0$, $t=50$.}
\label{fig4}
\end{figure*}

To evaluate the generalization performance of our model, we select the LaRE$^2$ model trained on the same datasets and consider it as the baseline method. Specifically, we train a classifier on each of the five selected subsets and then use each classifier to detect the other four different subsets. We conduct experiments using the validation set as the test set while keeping the dataset unchanged. We selected the time step \(t=0\) and \(t=50\), as they are relatively close to the end of the reverse process. The accuracies of different methods and detectors are shown in Figure~\ref{fig4} .

From the characteristics of the accuracy distribution, all approaches' accuracies along the diagonal are all close to $100\%$, indicating that when the generator is the same, the features learned by the classifier work well. However, when the classifier is applied to the images generated by other generators, the baseline performance showed significant variations. The classifier trained on diffusion model-based generated image subsets using TSG (\(t=0\)) shows great improvement in accuracy when tested on \textit{ADM}. This indicates that the features extracted using TSG demonstrate good generalization on diffusion model generated images. Besides, the model trained on \textit{SD V1.5}, and \textit{VQDM} can also work well on \textit{Biggan} subset. It shows that the classifier trained using our feature extraction method is not only effective for images generated by diffusion model-based generators but can also generalize to the \textit{Biggan}.

Since our method and baseline are designed based on the principles of diffusion models, the performance of the classifier impacts when trained on images generated by BigGAN. We will later explore ways to address this issue.
 
\subsection{Influence of Time Step $t$}
In TSG generation, we can control the time step \( t \) input to the neural network. This subsection discusses the impact of different time steps on feature extraction.

From Figure~\ref{fig4}, we can compare the impact of different time steps \( t \) on detection performance. Compared to \( t=0 \), as \( t \) increases, the amount of detailed information in the feature maps decreases. This results in the information content in the TSG images being insufficient, leading to poorer classification performance in distinguishing between real and fake images. Thus, the detection accuracy at \( t=50 \) slightly declines but remains, on average, higher than that of LaRE$^2$.

 \begin{figure*}[!ht]
\begin{center}
   \includegraphics[width=0.7\linewidth]{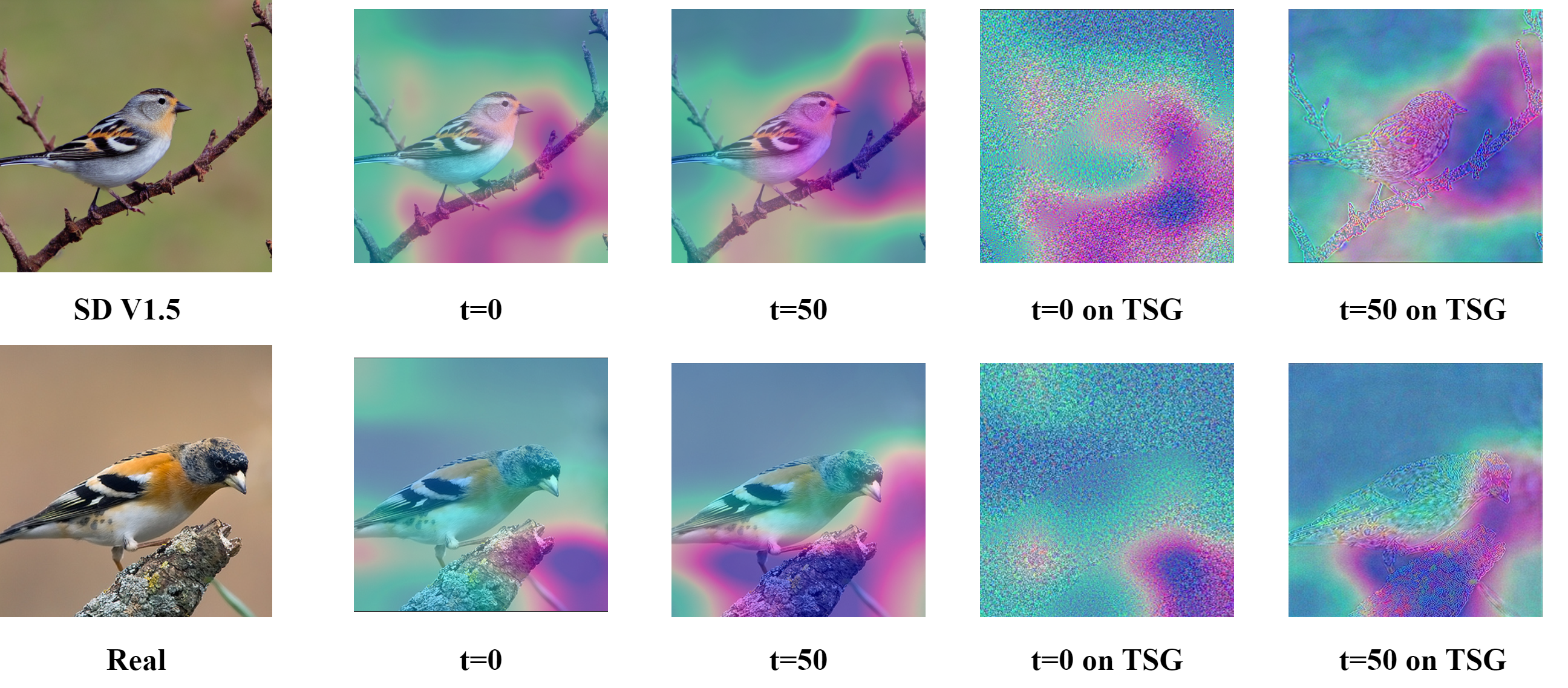}
\end{center}
   \caption{Using Grad-CAM heatmaps to demonstrate the part classifier relies for classification.}
\label{fig5}
\end{figure*}

\subsection{Performance Comparison with State of the Art}
Table 1 shows the accuracy comparisons on the selected dataset. From this, we can observe that our method significantly outperforms in accuracy, whether tested on diffusion model-based datasets or on \textit{BigGAN} generated dataset. The average accuracy shows an improvement of nearly 20 percentage compared to baseline LaRE$^2$.

\begin{table}[htbp]
    \centering
    \begin{tabular}{l|cc|c} 
        \toprule
        \multirow{2}{*}{Method} & \multicolumn{2}{c|}{Testing Subset} & \multirow{2}{*}{\makecell[c]{Avg\\ ACC.(\%)}} \\ 
        \cline{2-3}
        & \textit{Diff.-based} & \textit{BigGAN} & \\ 
        \midrule
        GramNet\cite{gradnet} & 65.0 & 62.4 & 63.7  \\ 
        DIRE\cite{dire} & 67.9 & 55.6 & 61.8 \\ 
        LaRE$^2$\cite{lare} & 78.8 & 72.4 & 75.6 \\ 
        \midrule
        \hb{TSG ($t=50$)} & \hb{87.5} & \hb{90.8} & \hb{89.2} \\ 
        \hr{TSG ($t=0$)}  & \hr{94.1} & \hr{95.6} & \hr{94.9}  \\ 
         \bottomrule
    \end{tabular}
    \caption{Accuracy comparisons on selected dataset. The best and second best results are hightled in \hr{red} and \hb{blue}. Diffusion based models include \textit{SD V1.5}, \textit{ADM}, \textit{VQDM}, and \textit{Wukong}. The accuracy of each subset is calculated using classifiers trained on five different selected subsets.}
    \label{tab1}
\end{table}

Subsequently, we demonstrate the speed advantages of our method through experiments. We conducted comparative tests to measure the time required by DIRE and TSG to generate 100 feature images in practical scenarios. The experiment was conducted on a single RTX A6000 GPU, and U-Net used in DIRE's reconstruction process is the same as the one used in TSG. The experimental parameters are configured to maximize memory utilization, shown in Table 2.

\begin{table}[htbp]
    \centering
    \begin{tabular}{l|ccc|c} 
        \toprule
        \multirow{2}{*}{Method} & \multicolumn{3}{c|}{Parameter} & \multirow{2}{*}{\makecell[c]{Time (s)}} \\ 
        \cline{2-4}
        & Batch Size & Sampling &Num& \\ 
        \midrule
        DIRE & 50 & “ddim20" & 100&277.1 \\ 
        TSG  & 5 & - - & 100& 26.3\\ 
        \bottomrule
    \end{tabular}
    \caption{Parameters and time required for DIRE and TSG.}
    \label{tab2}
\end{table}

As shown in the Table~\ref{tab2}, our method is nearly 10 times faster than DIRE. Since DIRE requires each image to go through 20 DDIM inversion steps and 20 denoising reconstruction steps, our method only needs to pass the image through the U-Net neural network once.

\subsection{Robustness Against Compression}
In studies focused on datasets for fake detection, there has been specific analysis \cite{jpeg} of how image size and quality impact classification results. We are particularly interested in whether the new features introduced by JPEG's lossy compression lead the classifier to learn some unintended features. Therefore, based on the quality of JPEG images, we selected images with a compression rate greater than 96 from the GenImage's subsets \textit{Glide}\cite{nichol2021glide}, \textit{SD V1.4}, and \textit{Midjourney} to create three new unbiased datasets. The specific composition of the dataset can be seen in Table~\ref{tab3}.

\begin{table}[htbp]
    \centering
    \begin{tabular}{l|cccc} 
        \toprule
        \multirow{2}{*}{Subset} & \multicolumn{2}{c}{Train} & \multicolumn{2}{c}{Testing }\\ 
        \cline{2-5}
        & Ai & Nature &Ai & Nature   \\ 
        \midrule
        \textit{Glide} & 113,085 & 113,085 &5000 & 5000 \\ 
        \textit{SD V1.4} &112,695  &112,695  &5000 & 5000  \\ 
        \textit{Midjourney} & 113,002 &113,002  &5000 & 5000  \\ 
         \bottomrule
    \end{tabular}
    \caption{The number of images in each of the three constructed Unbiased datasets.}
    \label{tab3}
\end{table}

\begin{figure}[t]
\begin{center}
   \includegraphics[width=0.55\linewidth]{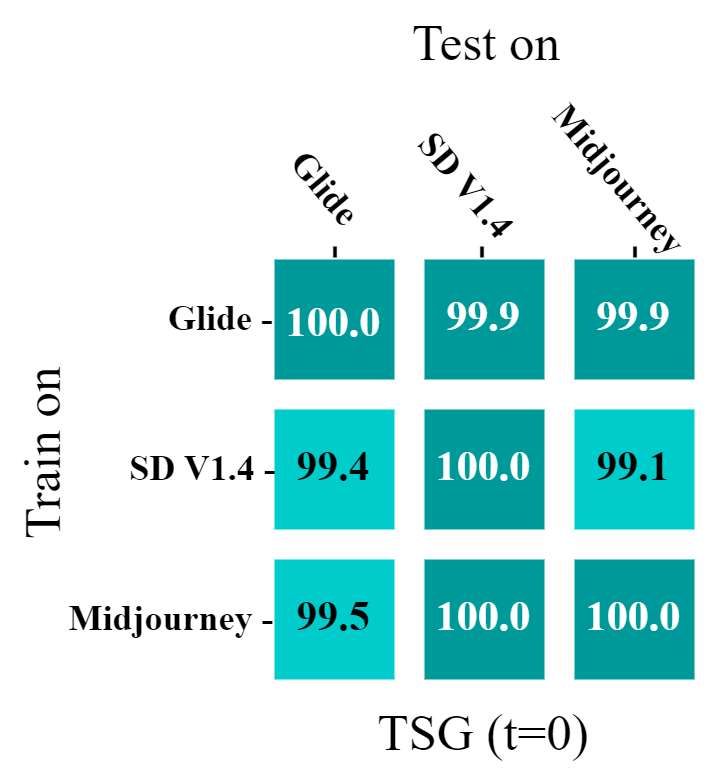} 
\end{center}
\caption{Cross-validation on an unbiased datasets.}
\label{fig6}
\end{figure}

Next, we perform cross-validation using these three unbiased datasets. The experiment result is shown in the Figure~\ref{fig6}.  From it, we observe that classifiers trained on the three datasets each demonstrate high accuracy during cross-validation. This suggests that our TSG model is not affected by noise introduced by  JPEG image compression.

\subsection{Grad-CAM Visualizations}
To further analyze how the classifier determines whether an image is generated based on features extracted by TSG, we use the Grad-CAM\cite{selvaraju2017grad} to visualize the ResNet-50 prediction process through heatmaps. 

The highlights in the heatmap represent contributions to the classification result. Figure~\ref{fig5} give the examples. From the whole-image perspective, the highlighted areas are extensive, indicating that the classifier does not make judgments based on just one or a few details. By comparing the heatmap distributions of the same image at \( t=0 \) and \( t=50 \), we can see that the areas contributing the most to the classification results are different. Moreover, at \( t=50 \), the classifier's attention is focused on the high-frequency details of the main object in the image, while at \( t=0 \), the heatmap shows a larger area of focus, which may due to the presence of more detailed features in the overall feature map at that time. This can also explains why TSG performs better in terms of accuracy at \( t=0 \) compared to \( t=50 \).

\subsection{Training on the Mixed dataset}
Our goal in the field of fake detection is to develop a fully generalizable detector that can accurately handle images generated by any type of generator. However, from Section 4.3, We observe a decrease in cross-validation accuracy for the BigGAN and diffusion series models. Therefore, we aim to enable the classifier to correctly classify images generated by two different types of generative models through TSG, while still using ResNet-50. We combined the training sets of the \textit{ADM}, \textit{SD V1.5}, and \textit{BigGAN} subsets and trained a classifier using TSG(t=0). The great performance of this new classifier is shown in Table 4. In which, \textit{Diff.-based} is the average accuracy mentioned in Section 4.5 and the \textit{Un-bias} is the average accuracy of three Un-bias subsets \textit{Glide}, \textit{SD V1.4}, and \textit{Midjourney}. 

\begin{table}[htbp]
    \centering
    \begin{tabular}{l|ccc} 
        \toprule
        Testing set& \textit{Diff.-based} & \textit{BigGAN}& \textit{Un-bias}   \\ 
        \midrule
        Accuracy(\%) & 100.0 &  100.0 & 100.0 \\ 
         \bottomrule
    \end{tabular}
    \caption{The performance of the classifier trained on the mixed dataset.}
    \label{tab4}
\end{table}

For further analysis, we use Grad-CAM to visualize  examples of misclassified images from the \textit{Wukong} and \textit{ADM} test sets by the classifier trained only on \textit{BigGAN}, and compared them with Grad-CAM visualizations of the same images identified by the classifier trained on the mixed dataset.
\begin{figure}[t]
\begin{center}
   \includegraphics[width=0.98\linewidth]{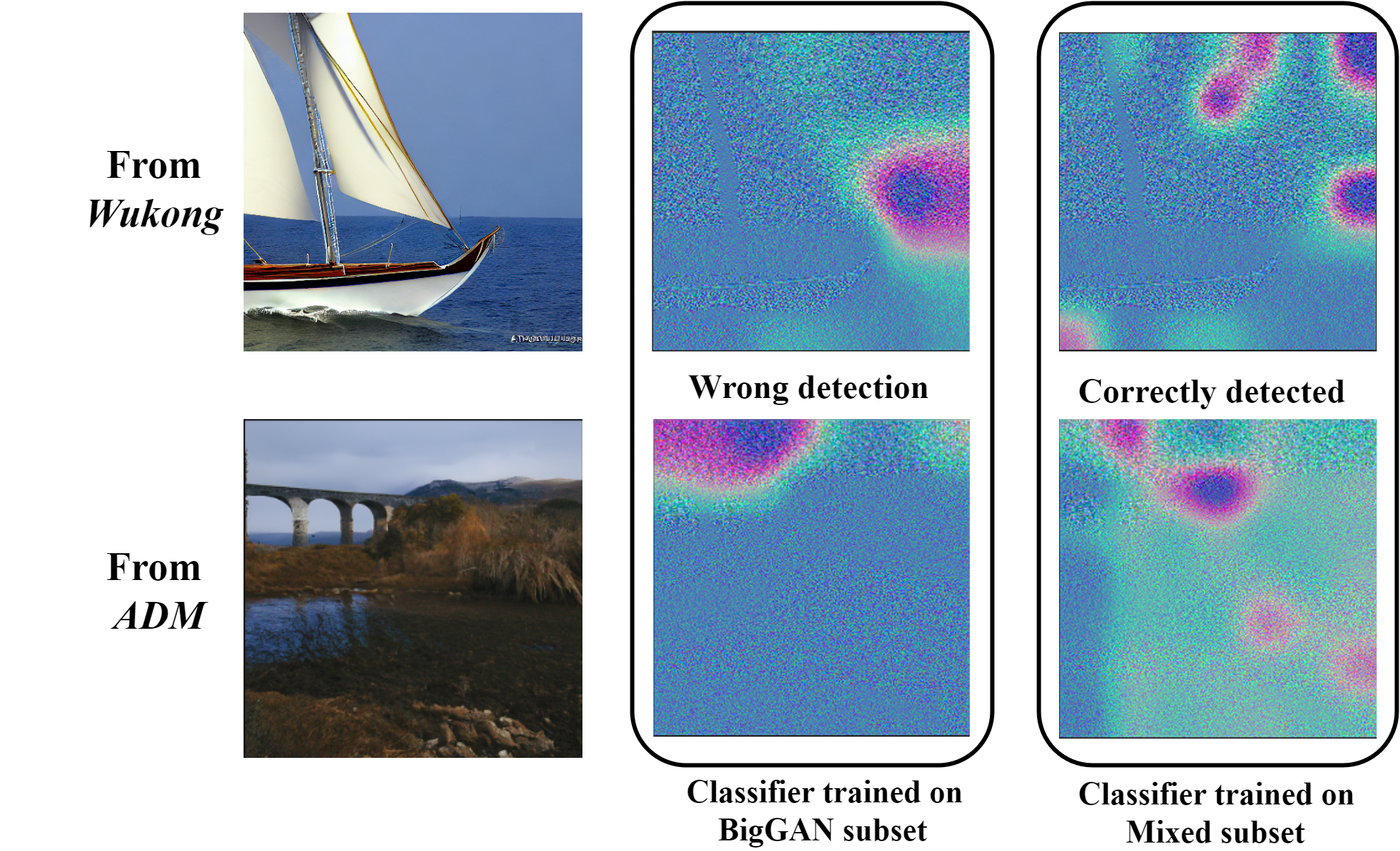} 
\end{center}
\caption{Grad-CAM visualizations for the same image at \( t=0 \): the misclassification by the classifier trained on \textit{BigGAN}, and the correct classification by the classifier trained on the mixed dataset.}
\label{fig7}
\end{figure}
Figure~\ref{fig7} shows various classification's Grad-CAM examples, illustrating that the classifier trained on the mixed dataset tends to extract details from the entire image, enabling more accurate predictions.

\section{Conclusion}
In this work, we innovatively addressed the problem of generated image detection by using the U-Net neural network from the pretrained diffusion model  as a feature extractor. This approach significantly improved detection accuracy while also accelerating the detection process. This method is more accurate than previous reconstruction-based detection models. In our experiments, we performed cross-validation, robustness testing, and heatmap analysis, all of which demonstrated the effectiveness and efficiency of our method. The proposed TSG is 19$\%$ better than the previous method LaRE$^2$ and the time required was only one-tenth of that for DIRE. 

{\small
\bibliographystyle{ieee_fullname}
\bibliography{egbib}

\begin{thebibliography}{10}\itemsep=-1pt

\bibitem{ddpm}
Jonathan Ho, Ajay Jain, and Pieter Abbeel.
\newblock Denoising diffusion probabilistic models.
\newblock {\em Advances in neural information processing systems}, 33:6840--6851, 2020.

\bibitem{ddim}
Jiaming Song, Chenlin Meng, and Stefano Ermon.
\newblock Denoising diffusion implicit models.
\newblock In {\em International Conference on Learning Representations}, 2021.

\bibitem{timesteptuner}
Mengfei Xia, Yujun Shen, Changsong Lei, Yu Zhou, Deli Zhao, Ran Yi, Wenping Wang, and Yong-Jin Liu.
\newblock Towards more accurate diffusion model acceleration with a timestep tuner.
\newblock In {\em Proceedings of the IEEE/CVF Conference on Computer Vision and Pattern Recognition}, pages 5736--5745, 2024.

\bibitem{attentionEfficiency}
Hongjie Wang, Difan Liu, Yan Kang, Yijun Li, Zhe Lin, Niraj~K Jha, and Yuchen Liu.
\newblock Attention-driven training-free efficiency enhancement of diffusion models.
\newblock In {\em Proceedings of the IEEE/CVF Conference on Computer Vision and Pattern Recognition}, pages 16080--16089, 2024.

\bibitem{diffussm}
Jing~Nathan Yan, Jiatao Gu, and Alexander~M Rush.
\newblock Diffusion models without attention.
\newblock In {\em Proceedings of the IEEE/CVF Conference on Computer Vision and Pattern Recognition}, pages 8239--8249, 2024.

\bibitem{zhang2024improving}
Huijie Zhang, Yifu Lu, Ismail Alkhouri, Saiprasad Ravishankar, Dogyoon Song, and Qing Qu.
\newblock Improving training efficiency of diffusion models via multi-stage framework and tailored multi-decoder architecture.
\newblock In {\em Proceedings of the IEEE/CVF Conference on Computer Vision and Pattern Recognition}, pages 7372--7381, 2024.

\bibitem{ADM}
Prafulla Dhariwal and Alexander Nichol.
\newblock Diffusion models beat gans on image synthesis.
\newblock In M. Ranzato, A. Beygelzimer, Y. Dauphin, P.S. Liang, and J.~Wortman Vaughan, editors, {\em Advances in Neural Information Processing Systems}, volume~34, pages 8780--8794. Curran Associates, Inc., 2021.

\bibitem{sohl2015deep}
Jascha Sohl-Dickstein, Eric Weiss, Niru Maheswaranathan, and Surya Ganguli.
\newblock Deep unsupervised learning using nonequilibrium thermodynamics.
\newblock In {\em International conference on machine learning}, pages 2256--2265. PMLR, 2015.

\bibitem{sd}
Robin Rombach, Andreas Blattmann, Dominik Lorenz, Patrick Esser, and Bj\"orn Ommer.
\newblock High-resolution image synthesis with latent diffusion models.
\newblock In {\em Proceedings of the IEEE/CVF Conference on Computer Vision and Pattern Recognition (CVPR)}, pages 10684--10695, June 2022.

\bibitem{nichol2021improved}
Alexander~Quinn Nichol and Prafulla Dhariwal.
\newblock Improved denoising diffusion probabilistic models.
\newblock In {\em International conference on machine learning}, pages 8162--8171. PMLR, 2021.

\bibitem{ho2022classifier}
Jonathan Ho and Tim Salimans.
\newblock Classifier-free diffusion guidance.
\newblock {\em arXiv preprint arXiv:2207.12598}, 2022.

\bibitem{videodiff}
Jonathan Ho, Tim Salimans, Alexey Gritsenko, William Chan, Mohammad Norouzi, and David~J Fleet.
\newblock Video diffusion models.
\newblock {\em Advances in Neural Information Processing Systems}, 35:8633--8646, 2022.

\bibitem{controlnetplus}
Ming Li, Taojiannan Yang, Huafeng Kuang, Jie Wu, Zhaoning Wang, Xuefeng Xiao, and Chen Chen.
\newblock Controlnet $++$: Improving conditional controls with efficient consistency feedback.
\newblock In {\em European Conference on Computer Vision}, pages 129--147. Springer, 2025.

\bibitem{peng2024controlnext}
Bohao Peng, Jian Wang, Yuechen Zhang, Wenbo Li, Ming-Chang Yang, and Jiaya Jia.
\newblock Controlnext: Powerful and efficient control for image and video generation.
\newblock {\em arXiv preprint arXiv:2408.06070}, 2024.

\bibitem{imagic}
Bahjat Kawar, Shiran Zada, Oran Lang, Omer Tov, Huiwen Chang, Tali Dekel, Inbar Mosseri, and Michal Irani.
\newblock Imagic: Text-based real image editing with diffusion models.
\newblock In {\em Proceedings of the IEEE/CVF Conference on Computer Vision and Pattern Recognition}, pages 6007--6017, 2023.

\bibitem{brooks2023instructpix2pix}
Tim Brooks, Aleksander Holynski, and Alexei~A Efros.
\newblock Instructpix2pix: Learning to follow image editing instructions.
\newblock In {\em Proceedings of the IEEE/CVF Conference on Computer Vision and Pattern Recognition}, pages 18392--18402, 2023.

\bibitem{juefei2022countering}
Felix Juefei-Xu, Run Wang, Yihao Huang, Qing Guo, Lei Ma, and Yang Liu.
\newblock Countering malicious deepfakes: Survey, battleground, and horizon.
\newblock {\em International journal of computer vision}, 130(7):1678--1734, 2022.

\bibitem{dire}
Zhendong Wang, Jianmin Bao, Wengang Zhou, Weilun Wang, Hezhen Hu, Hong Chen, and Houqiang Li.
\newblock Dire for diffusion-generated image detection.
\newblock In {\em Proceedings of the IEEE/CVF International Conference on Computer Vision (ICCV)}, pages 22445--22455, October 2023.

\bibitem{lare}
Yunpeng Luo, Junlong Du, Ke Yan, and Shouhong Ding.
\newblock Lare{\textasciicircum}2: Latent reconstruction error based method for diffusion-generated image detection.
\newblock In {\em Proceedings of the IEEE/CVF Conference on Computer Vision and Pattern Recognition (CVPR)}, pages 17006--17015, June 2024.

\bibitem{zhu2024genimage}
Mingjian Zhu, Hanting Chen, Qiangyu Yan, Xudong Huang, Guanyu Lin, Wei Li, Zhijun Tu, Hailin Hu, Jie Hu, and Yunhe Wang.
\newblock Genimage: A million-scale benchmark for detecting ai-generated image.
\newblock {\em Advances in Neural Information Processing Systems}, 36, 2024.

\bibitem{gan}
Ian Goodfellow, Jean Pouget-Abadie, Mehdi Mirza, Bing Xu, David Warde-Farley, Sherjil Ozair, Aaron Courville, and Yoshua Bengio.
\newblock Generative adversarial nets.
\newblock In Z. Ghahramani, M. Welling, C. Cortes, N. Lawrence, and K.Q. Weinberger, editors, {\em Advances in Neural Information Processing Systems}, volume~27. Curran Associates, Inc., 2014.

\bibitem{vae}
Diederik~P Kingma.
\newblock Auto-encoding variational bayes.
\newblock {\em arXiv preprint arXiv:1312.6114}, 2013.

\bibitem{pixel}
Junsong Chen, Jincheng YU, Chongjian GE, Lewei Yao, Enze Xie, Zhongdao Wang, James Kwok, Ping Luo, Huchuan Lu, and Zhenguo Li.
\newblock Pixart-\${\textbackslash}alpha\$: Fast training of diffusion transformer for photorealistic text-to-image synthesis.
\newblock In {\em The Twelfth International Conference on Learning Representations}, 2024.

\bibitem{wang2020cnn}
Sheng-Yu Wang, Oliver Wang, Richard Zhang, Andrew Owens, and Alexei~A Efros.
\newblock Cnn-generated images are surprisingly easy to spot... for now.
\newblock In {\em Proceedings of the IEEE/CVF conference on computer vision and pattern recognition}, pages 8695--8704, 2020.

\bibitem{chai2020makes}
Lucy Chai, David Bau, Ser-Nam Lim, and Phillip Isola.
\newblock What makes fake images detectable? understanding properties that generalize.
\newblock In {\em Computer Vision--ECCV 2020: 16th European Conference, Glasgow, UK, August 23--28, 2020, Proceedings, Part XXVI 16}, pages 103--120. Springer, 2020.

\bibitem{face1}
Yurika~Fujinami Hiroshi~Watanabe.
\newblock Adversarial level of face images generated by prompt-based image coding in face recognition system.
\newblock In {\em IEEE Global Conference on Consumer Electronics (GCCE2024).}, pages 332--333, 2024.

\bibitem{liang2024poisoned}
Jiawei Liang, Siyuan Liang, Aishan Liu, Xiaojun Jia, Junhao Kuang, and Xiaochun Cao.
\newblock Poisoned forgery face: Towards backdoor attacks on face forgery detection.
\newblock {\em arXiv preprint arXiv:2402.11473}, 2024.

\bibitem{zhang2022improving}
Mingxu Zhang, Hongxia Wang, Peisong He, Asad Malik, and Hanqing Liu.
\newblock Improving gan-generated image detection generalization using unsupervised domain adaptation.
\newblock In {\em 2022 IEEE International Conference on Multimedia and Expo (ICME)}, pages 1--6. IEEE, 2022.

\bibitem{wissmann2024whodunit}
Alexander Wi{\ss}mann, Steffen Zeiler, Robert~M Nickel, and Dorothea Kolossa.
\newblock Whodunit: Detection and attribution of synthetic images by leveraging model-specific fingerprints.
\newblock In {\em Proceedings of the 3rd ACM International Workshop on Multimedia AI against Disinformation}, pages 65--72, 2024.

\bibitem{he2024rigid}
Zhiyuan He, Pin-Yu Chen, and Tsung-Yi Ho.
\newblock Rigid: A training-free and model-agnostic framework for robust ai-generated image detection.
\newblock {\em arXiv preprint arXiv:2405.20112}, 2024.

\bibitem{wu2023}
Haiwei Wu, Jiantao Zhou, and Shile Zhang.
\newblock Generalizable synthetic image detection via language-guided contrastive learning.
\newblock {\em arXiv preprint arXiv:2305.13800}, 2023.

\bibitem{clip}
Alec Radford, Jong~Wook Kim, Chris Hallacy, Aditya Ramesh, Gabriel Goh, Sandhini Agarwal, Girish Sastry, Amanda Askell, Pamela Mishkin, Jack Clark, et~al.
\newblock Learning transferable visual models from natural language supervision.
\newblock In {\em International conference on machine learning}, pages 8748--8763. PMLR, 2021.

\bibitem{Cazenavette_2024}
George Cazenavette, Avneesh Sud, Thomas Leung, and Ben Usman.
\newblock Fakeinversion: Learning to detect images from unseen text-to-image models by inverting stable diffusion.
\newblock In {\em Proceedings of the IEEE/CVF Conference on Computer Vision and Pattern Recognition (CVPR)}, pages 10759--10769, June 2024.

\bibitem{Tan_2023}
Chuangchuang Tan, Yao Zhao, Shikui Wei, Guanghua Gu, and Yunchao Wei.
\newblock Learning on gradients: Generalized artifacts representation for gan-generated images detection.
\newblock In {\em Proceedings of the IEEE/CVF Conference on Computer Vision and Pattern Recognition (CVPR)}, pages 12105--12114, June 2023.

\bibitem{resnet}
Kaiming He, Xiangyu Zhang, Shaoqing Ren, and Jian Sun.
\newblock Deep residual learning for image recognition.
\newblock In {\em Proceedings of the IEEE Conference on Computer Vision and Pattern Recognition (CVPR)}, June 2016.

\bibitem{biggan}
Andrew Brock, Jeff Donahue, and Karen Simonyan.
\newblock Large scale {GAN} training for high fidelity natural image synthesis.
\newblock In {\em International Conference on Learning Representations}, 2019.

\bibitem{gu2022vector}
Shuyang Gu, Dong Chen, Jianmin Bao, Fang Wen, Bo Zhang, Dongdong Chen, Lu Yuan, and Baining Guo.
\newblock Vector quantized diffusion model for text-to-image synthesis.
\newblock In {\em Proceedings of the IEEE/CVF conference on computer vision and pattern recognition}, pages 10696--10706, 2022.

\bibitem{gu2022wukong}
Jiaxi Gu, Xiaojun Meng, Guansong Lu, Lu Hou, Niu Minzhe, Xiaodan Liang, Lewei Yao, Runhui Huang, Wei Zhang, Xin Jiang, et~al.
\newblock Wukong: A 100 million large-scale chinese cross-modal pre-training benchmark.
\newblock {\em Advances in Neural Information Processing Systems}, 35:26418--26431, 2022.

\bibitem{gradnet}
Zhengzhe Liu, Xiaojuan Qi, and Philip~H.S. Torr.
\newblock Global texture enhancement for fake face detection in the wild.
\newblock In {\em Proceedings of the IEEE/CVF Conference on Computer Vision and Pattern Recognition (CVPR)}, June 2020.

\bibitem{jpeg}
Patrick Grommelt, Louis Weiss, Franz-Josef Pfreundt, and Janis Keuper.
\newblock Fake or jpeg? revealing common biases in generated image detection datasets.
\newblock {\em arXiv preprint arXiv:2403.17608}, 2024.

\bibitem{nichol2021glide}
Alex Nichol, Prafulla Dhariwal, Aditya Ramesh, Pranav Shyam, Pamela Mishkin, Bob McGrew, Ilya Sutskever, and Mark Chen.
\newblock Glide: Towards photorealistic image generation and editing with text-guided diffusion models.
\newblock {\em arXiv preprint arXiv:2112.10741}, 2021.

\bibitem{selvaraju2017grad}
Ramprasaath~R Selvaraju, Michael Cogswell, Abhishek Das, Ramakrishna Vedantam, Devi Parikh, and Dhruv Batra.
\newblock Grad-cam: Visual explanations from deep networks via gradient-based localization.
\newblock In {\em Proceedings of the IEEE international conference on computer vision}, pages 618--626, 2017.

\end{thebibliography}
}

\end{document}